\PassOptionsToPackage{prologue,dvipsnames}{xcolor}

\documentclass[10pt,twocolumn,letterpaper]{article}

\usepackage[accsupp]{axessibility}  

\usepackage{iccv}
\usepackage{times}
\usepackage{epsfig}
\usepackage{graphicx}
\usepackage{amsmath}
\usepackage{amssymb}
\usepackage{subcaption}
\usepackage{mathabx,graphicx}
\usepackage{amssymb}
\usepackage{booktabs}
\usepackage{multirow}
\usepackage{enumitem}
\usepackage{mathabx,graphicx}
\usepackage{algpseudocode}
\usepackage{algorithm}
\usepackage{colortbl}
\usepackage[cmyk]{xcolor}
\usepackage{array}
\definecolor{c1}{HTML}{f0f0f4}
\definecolor{green1}{RGB}{84,139,84}

\usepackage[dvipsnames]{xcolor}
\def\eg{\emph{e.g.}} 

\def\ie{\emph{i.e. }}

\def\etal{\emph{et al. }}

\usepackage[pagebackref=true,breaklinks=true,letterpaper=true,colorlinks,bookmarks=false]{hyperref}
\usepackage[linewidth=1pt]{mdframed}
\iccvfinalcopy 

\def\iccvPaperID{3151} 

\ificcvfinal\fi

\begin{document}

\title{March in Chat: Interactive Prompting for\\ Remote Embodied Referring Expression}

\author{
Yanyuan Qiao$^{1}$ \quad Yuankai Qi$^{1}$ \quad Zheng Yu$^{1}$  \quad Jing Liu$^{2}$ $^{3}$ \quad Qi Wu$^{1}$\thanks{Corresponding author}\\
$^1$Australian Institute for Machine Learning, The University of Adelaide\\$^2$Institute of Automation, Chinese Academy of Sciences\\
~~ $^3$School of Artificial Intelligence, University of Chinese Academy of Sciences\\
{\tt\small \{yanyuan.qiao,zheng.yu,qi.wu01\}@adelaide.edu.au, qykshr@gmail.com,jliu@nlpr.ia.ac.cn}\\
}

\maketitle
\ificcvfinal\thispagestyle{empty}\fi

\begin{abstract}
Many Vision-and-Language Navigation (VLN) tasks have been proposed in recent years, from room-based to object-based and indoor to outdoor.
The REVERIE (Remote Embodied Referring Expression) is interesting since it only provides high-level instructions to the agent, which are closer to human commands in practice. Nevertheless, this poses more challenges than other VLN tasks since it requires agents to infer a navigation plan only based on a short instruction. Large Language Models (LLMs) show great potential in robot action planning by providing proper prompts.
Still, this strategy has not been explored under the REVERIE settings. There are several new challenges. For example, the LLM should be environment-aware so that the navigation plan can be adjusted based on the current visual observation. Moreover, the LLM planned actions should be adaptable to the much larger and more complex REVERIE environment. This paper proposes a March-in-Chat (MiC) model that can talk to the LLM on the fly and plan dynamically based on a newly proposed Room-and-Object Aware Scene Perceiver (ROASP). Our MiC model outperforms the previous state-of-the-art by large margins by SPL and RGSPL metrics on the REVERIE benchmark. The source code is available at \href{https://github.com/YanyuanQiao/MiC}{https://github.com/YanyuanQiao/MiC}

\end{abstract}

\section{Introduction}
\label{sec:intro}


Vision-and-Language Navigation (VLN), which lies at the intersection of computer vision, natural language processing and robotics, has aroused great attention from research communities in the past few years. Given instructions in natural language, the VLN agent should navigate to the target location based on the dynamic observations in the 3D simulated environments. Since VLN has great potential in real-world applications such as domestic assistant robots, a large amount of specific VLN tasks have been proposed, including R2R~\cite{r2r} and RxR~\cite{rxr} that ask the agent to navigate from one room to another in a photo-realistic environment according to the fine-grained instruction, NDH~\cite{ndh} provides detailed dialogues which imply the instruction, TouchDown~\cite{touchdown} extends the task into an outdoor environment, REVERIE~\cite{reverie} and SOON~\cite{soon} that additionally
require the agent's ability of remote object grounding and ALFRED~\cite{alfred} that asks the agent to interact with the target object in  a single room of the synthetic environment.

\begin{figure}[t]
\begin{center}
\includegraphics[width=1\linewidth]{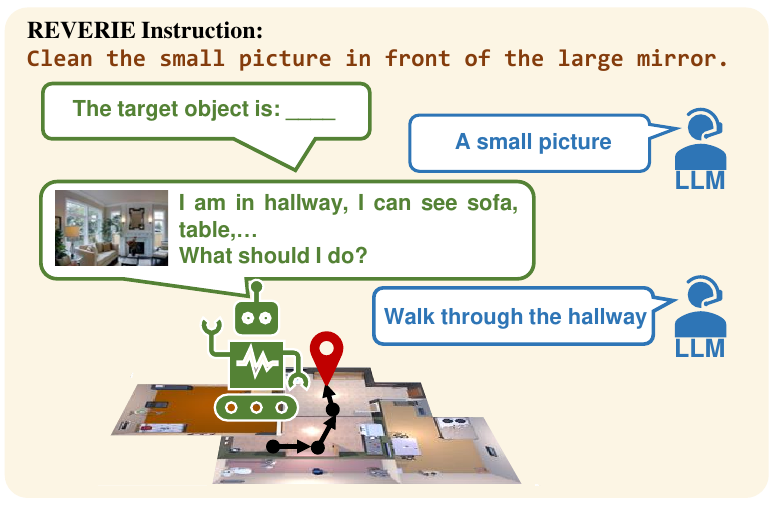}
\end{center}
\vspace{-15pt}
   \caption{Our March-in-Chat (MiC) model is talking to a Large Language Model (LLM) to generate navigation plans on the fly, with the REVERIE instruction and the dynamic room-and-object information as inputs.}
\vspace{-15pt}
\label{fig:intro}
\end{figure}

Most of these VLN tasks provide detailed step-by-step instructions to the agent, such as ``Go up the stairs and then walk the length of the couch. Walk past the dining area and into the kitchen. Stop in front of the refrigerator.'' in R2R. Although detailed instructions can help the agent better achieve the navigation goal in the simulated environments, it has a big gap towards real applications where human beings tend to give coarse-grained high-level instructions such as ``Go to the refrigerator on the second floor''. Contrary to other tasks, the Remote Embodied Referring Expression (REVERIE) task is more likely to empower the real-world applications of VLN, of which the instructions are closer to those in practice, such as ``Empty the washing machine on level one''. Such high-level instruction is more challenging for VLN agents since it requires them to be more competent in perceiving the surrounding environment and the navigation progress and correspondingly making reasonable plans for the next steps.

Recently, Large Language Models (LLMs) that internalize a wealth of commonsense knowledge show great potential in action planning for some embodied tasks with the help of suitable in-context learning. However, previous works mainly utilize LLMs to plan atomic actions of object manipulation in a very limited space with simple scenes. These predefined atomic actions can be easily planned well by the LLMs planners with a unified template. Different from these embodied tasks, REVERIE requires large-area exploration from one room to another, which is complex in the layout of rooms and scenes with diverse objects. 

In this work, to adapt LLMs as the planner for REVERIE with the ability of comprehensive scene perception, we propose a novel model named \textit{March in Chat} (MiC), which enables the LLM as an environment-aware instruction planner through on-the-fly dialogues between the agent and the LLM as Fig.~\ref{fig:intro} shows. Specifically, the agent is initially situated at the starting position given a high-level coarse-grained REVERIE instruction. First, a Goal-Oriented Static Planning (\textbf{GOSP}) module queries the LLM to point out the target object and infer where the thing may be by using the rich world knowledge internalized in the LLM. Secondly, the agent's Room-and-Object Aware Scene Perceiver (\textbf{ROASP}) describes the current observation and asks the LLM to generate step-by-step fine-grained planning for the next navigation steps. Then, if the ROASP finds the room has changed, the LLM is queried again by the Scene-Oriented Dynamic Planning (\textbf{SODP}) module to generate a new fine-grained step-by-step planning, which will be concatenated with all previous responses from the LLM. The agent will march under the guidance of such interactive prompting until the task is finished.

To evaluate our proposed MiC, we conduct experiments on the REVERIE benchmark. Our MiC achieves a new state-of-the-art performance in all metrics on REVERIE val unseen set and REVERIE test unseen set. Mainly, MiC obtains 41.97\% on the primary navigation metric of SPL and 26.17\% on the major object grounding metric of RGSPL on test split, which is at least 3.09\% and 3.49\% higher than the previous SoTA results. We also conduct ablation studies to validate the contributions of different components in MiC and the effect of scene-aware perception in dynamic planning generation. These promising results demonstrate the effectiveness of our proposed MiC.

In summary, we make the following contributions:
\vspace{-5pt}
\begin{itemize}
    \item We propose a novel March-in-Chat (MiC) model, which lets the REVERIE agent talk with an LLM on the fly to make plans for the next few steps.
    \vspace{-5pt}
    \item Two planning modules, namely Goal-Oriented Static Planning (GOSP) module, and Scene-Oriented Dynamic Planning (SODP) module, and one Room-and-Object Aware Scene Perceiver (ROASP) module, are proposed.
    \vspace{-5pt}
    \item Extensive quantitative and qualitative experiments are conducted on REVERIE to validate the effectiveness of our method.
\end{itemize}

\section{Related work}
\label{sec:related_work}

\noindent\textbf{Vision-and-Language Navigation}
Vision-and-Language Navigation (VLN) has attracted increasing attention in recent years, and many specific VLN tasks have been proposed~\cite{touchdown,r4r,hanna,reverie,qiao2023hop+,ndh}.
Anderson~\etal\cite{r2r} proposes the first VLN benchmark, Room-to-Room (R2R), which requires an agent to navigate from one room to another in a house, according to a detailed natural language step-by-step instruction in a photo-realistic environment. Later, Room-across-Room (RxR)~\cite{rxr} was proposed with longer and more detailed multilingual instructions. Both these two tasks give fine-grained instructions, which makes it easier to navigate to the target location. NDH~\cite{ndh} extends the navigation instruction to the dialogue form, and TouchDown~\cite{touchdown} extends the environments to outdoor.
REVERIE~\cite{reverie} and SOON~\cite{soon} are proposed for remote object localization, which requires an agent not only to navigate to the target location, but also to specify the object to interact with.
The difference between REVERIE and SOON is that REVERIE uses short concise instructions (\eg, ``bring me the red cup from the kitchen.'') while SOON employs long detailed instructions (\eg, ``I want to find a cylindrical, metallic and tall lamp, which is set in the bright living room. The lamp is on the cabinet which is on the left of the television and next to the window. The living room is on ...'' ).

Among the aforementioned VLN tasks, the instructions of REVERIE are closer to what we would say to an intelligent domestic robot in daily life in terms of the instruction length and logic, which is usually short and concise. 
However, most existing methods~\cite{an2022bevbert,Chen_2022_DUET,he2021landmark,recurrent,orist,hop} are usually designed for the VLN tasks where  detailed step-by-step instructions are used, thus they do not perform well on REVERIE.
In this work, we specifically develop a method for REVERIE. Inspired by the fact of LLMs that implicitly internalize rich knowledge in action planning, we propose to exploit LLM as a fine-grained planner to generate detailed navigation plans from the concise instructions of REVERIE to improve navigation success.

\begin{figure*}[!t]
\begin{center}
\includegraphics[width=0.99\linewidth]{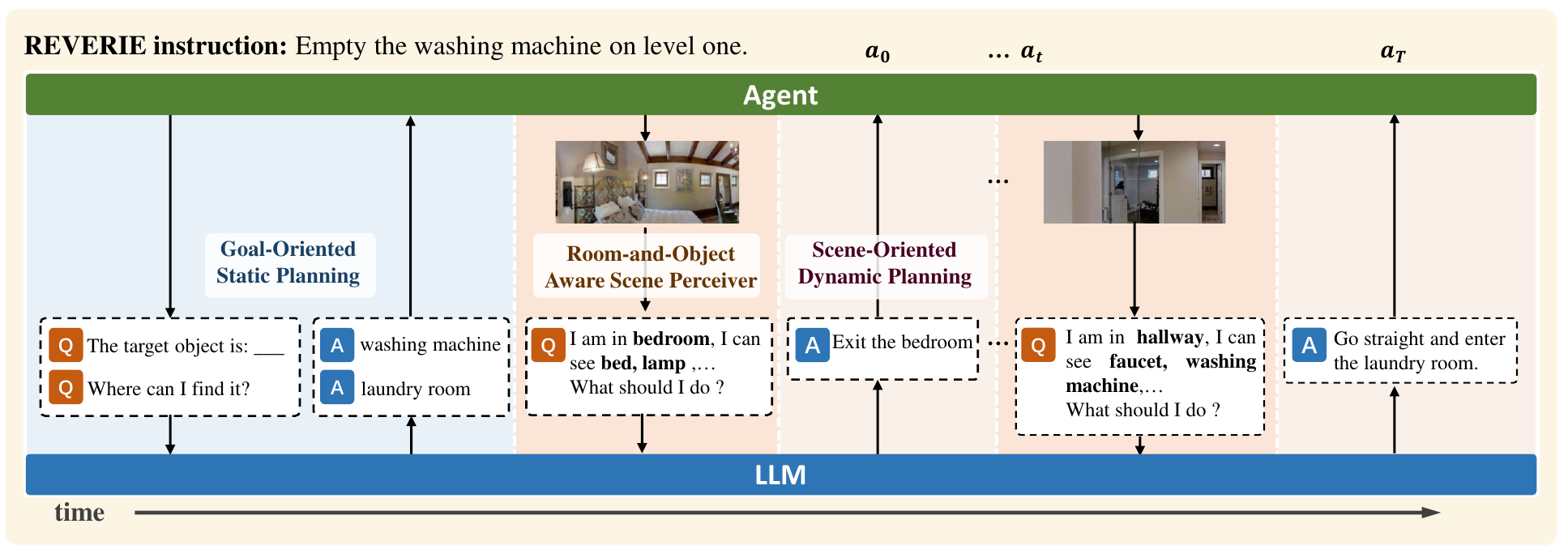}
\vspace{-17pt}
\end{center}
   \caption{Overview of our March-in-Chat model. The program runs along the vertical arrows from left to right, progressing with the time flow. Our model first performs Goal-Oriented Static Planning (GOSP, Sec.~\ref{subsubsec:GOSP}) to reason the target object and its possible lying room; then the Room-and-Object Aware Scene Perceiver (ROASP, Sec.~\ref{subsec:monitor}) perceives what room type the agent currently stands in  and what prominent objects can be seen; these information are used by the Scene-Oriented Dynamic Planning module (SODP, Sec.~\ref{subsubsec:SODP}) to generate a detailed instruction to execute. GOSP just runs once, and we repeat ROASP and SODP until the agent chooses to stop or reaches the maximum steps.}
\label{fig:overview}
\vspace{-10pt}
\end{figure*}

\noindent\textbf{LLMs as Embodied Planner} 
Benefiting from the rise of LLMs, recent works~\cite{huang2022language,progprompt} have explored the use of LLMs in task planning for various embodied tasks.
Huang~\etal\cite{huang2022language}  propose to utilize the frozen LLMs (\eg, GPT-2~\cite{gpt2}, GPT-3~\cite{gpt3} and Codex~\cite{codex}) to plan actions for the embodied agent with in-context learning~\cite{gpt3}.
SayCan~\cite{saycan} translates a high-level instruction into a list of candidate low-level actions with a probability, which is then multiplied by a value function for action prediction. 
These two LLM planners are static, which only generate action plans at the beginning of a task. By contrast, Huang~\etal~\cite{huang2022inner} propose to introduce the feedback of action progress, detected objects and human assistance into the LLM planner to re-plan  atomic actions. One concurrent work by Song \etal\cite{song2022llm} injects the detected objects to re-generate  high-level plans with a fixed program pattern for the ALFRED~\cite{alfred} task.

However, these above-mentioned methods mainly concentrate on planning atomic actions for object manipulation in a very limited space with simple scenes.
By contrast, REVERIE has plenty of much larger and more complicated environments: 90 multi-layer buildings of various styles (\eg, office, home, gym, to name a few).
To handle the complex scenarios of REVERIE, we propose a Room-and-Object Aware Scene
Perceiver module that helps the LLM planner dynamically interact with the environment in the form of a natural language dialogue.
\section{Problem Setup}
\label{sec:pre}

In the REVERIE task, given a concise and high-level instruction referring to a remote object, the agent is expected to navigate to the goal location and identify the target object in previously unseen environments.
The environment is defined as an undirected graph $\mathcal{G}=\{\mathcal{V}, \mathcal{E}\}$, where $\mathcal{V}=\{V_i\}_{i=1}^{K}$ denotes $K$ navigable nodes, and $\mathcal{E}$ denotes connectivity edges.
The agent is first placed in a starting node with the initial state $s_0$ and perceives a panorama $\mathcal R_t$ as the visual observation at each time step $t$.
The panorama $\mathcal R_t$ is split into $n$ single view images as $\mathcal R_t=\left \{ r_i \right \}_{i=1}^{n}$. Each single view image $r_i$ is represented by an image feature vector and an orientation feature vector. 
In addition, the object features $\mathcal O_t =\left \{ o_i \right \}_{i=1}^{m}$ of $m$ objects are extracted from the panorama view using the annotated object bounding boxes or object detectors.
Then, the agent makes a sequence of actions $\langle{a_0},...{a_T}\rangle$ to reach the target location, where each action is achieved by choosing a navigable node from the candidate list.
The agent navigates in the environment until the target object is grounded or the agent reaches the pre-defined maximum trajectory length.

\vspace{-3pt}
\section{Method}
\label{sec:methods}

\begin{figure*}[!t]
\centering
\includegraphics[width=0.85\linewidth]{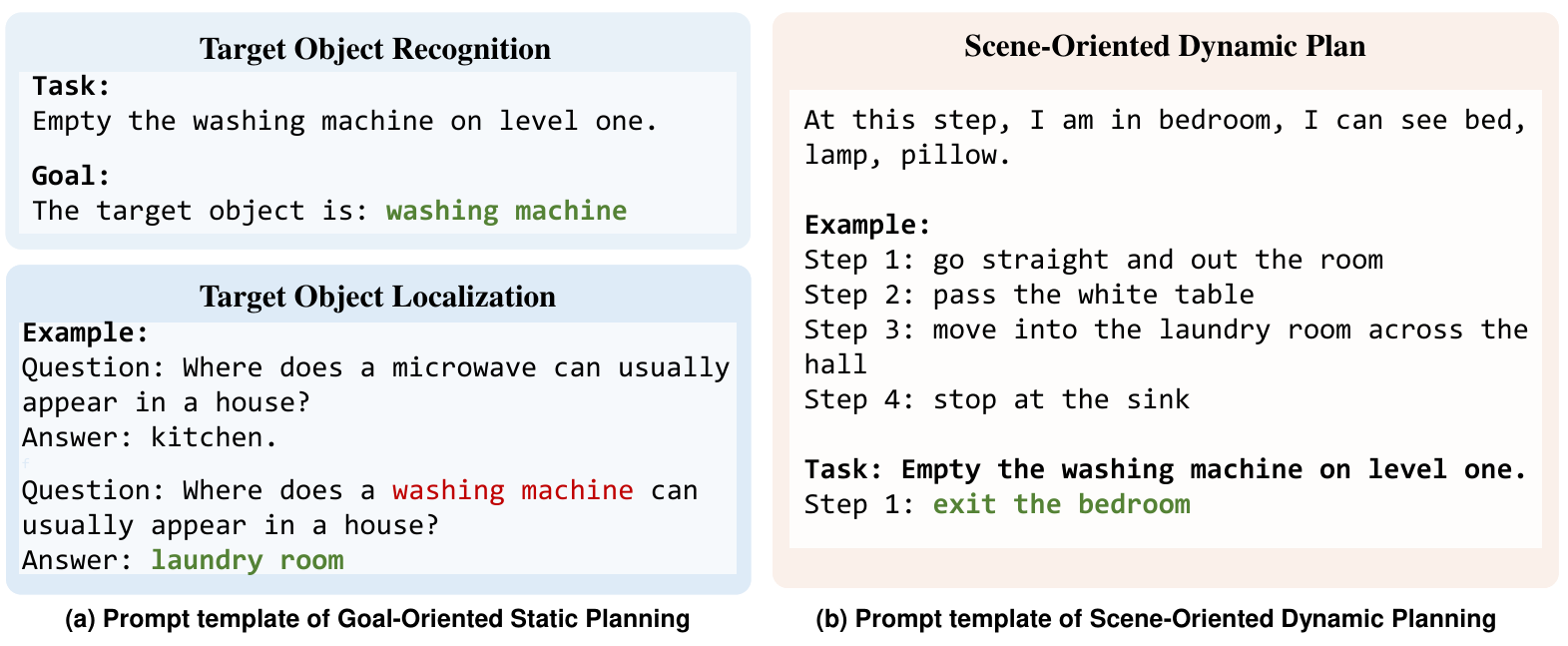}
\vspace{-8pt}
   \caption{Examples of prompting templates for Goal-Oriented Static Planning (a) and Scene-Oriented Dynamic Planning (b). Outputs are marked with \textbf{\color{green1}{green}} color. And the \textbf{\textcolor{red}{red}} denotes the predicted object from Target Object Recognition.}
   \vspace{-8pt}
\label{fig:prompt}
\end{figure*}

As illustrated in Fig.~\ref{fig:overview}, when initially situated at the starting position and given a concise high-level instruction such as ``Empty the washing machine on level one'', the agent first queries an LLM with the \textbf{GOSP} module (Sec. \ref{subsubsec:GOSP}) to find out the target object  ``\texttt{washing machine}'' in the instruction and reason out the potential location ``\texttt{laundry room}'' by using the world knowledge implied in an LLM. 
Then the \textbf{ROASP} module (Sec. \ref{subsec:monitor}) extracts the room type and visible objects from the current visual observation to obtain the environmental feedback.
With the description of the scene perception, the LLM is queried again by the \textbf{SODP} (Sec. \ref{subsubsec:SODP}) to generate the next step instruction, which is used to guide the agent to navigate to the target object in the room. 
The baseline agent is based on HM3D-DUET~\cite{Chen_2022_HM3D_AutoVLN}, more details can be found in the supplementary material. 

\vspace{-8pt}
\subsection{Planning with World Knowledge from LLMs}
\label{subsec:world}
This section illustrates how we utilize in-context learning (ICL)~\cite{gpt3} to acquire world knowledge from the LLMs for planning. 
We first briefly introduce in-context learning.
Then, we elaborate the 
Goal-Oriented Static Planning (GOSP) and Scene-Oriented Dynamic Planning (SODP). 
Last, we show the demonstration selection process. 

\noindent\textbf{Preliminary: In-context Learning for Planning.}
In-context learning (ICL) is a paradigm that lets LLMs directly make predictions based on a natural language context without gradient updates~\cite{gpt3}.
Specifically, under the setting of in-context learning, an LLM is fed a ``\texttt{prompt}'' that usually contains a task description and several demonstrations, and then the LLM  generates the required outputs. 
Both the prompt template and the choice of demonstration examples have an impact on how well ICL performs. 
In this work, we use two different ICL settings to generate different navigation plans, \ie the GOSP and SODP module.

The GOSP aims to identify the target object and infer the target location by arousing the world knowledge contained in an LLM through appropriate prompts. A fixed demonstration example is used for GOSP.
While the SODP aims to generate step-by-step planning instructions after observing the dynamic scenes from the environment, which is more complicated than the former. 
To better generate plannings, we dynamically select the most suitable demonstration examples for SODP and incorporate the environmental feedback as prompts for interactive planning. 

\vspace{-8pt}
\subsubsection{Goal-Oriented Static Planning (GOSP)} 
\label{subsubsec:GOSP}
\vspace{-4pt}
Given a high-level concise instruction, such as ``Empty the washing machine on level one'', 
an LLM is first asked to generate a goal-oriented static planning instruction: ``Goal: The target object is a washing machine. It is usually in a laundry room'', 
which emphasizes the target object and points out where the target object may lie. 
As shown in Fig.~\ref{fig:prompt}(a), the planning generation mainly consists of two sub-tasks: target object recognition and target object localization, which can be achieved by providing specifically designed prompts for the LLM. 
To this end, we design the prompts for the former sub-task in the form: ``\textbf{Task:} Empty the washing machine on level one. \textbf{Goal:} The target object is: ''. Then the LLM will generate a corresponding answer: ``\texttt{washing machine}''. 
By contrast, the latter sub-task, reasoning out the target location, is more complex because it requires more suitable prompts to arouse the internalized world knowledge in the LLM. 
To address this problem, we utilize a fixed demonstration example for the LLM, and design the prompts for the latter sub-task  in the form: ``\textbf{Example:} Question: Where does a microwave can usually appear in a house?  Answer: kitchen.  Question: Where does a \texttt{washing machine} can usually appear in a house?  Answer: ''. 
Given such prompts, the LLM will generate the corresponding answer ``\texttt{laundry room}''. With these answers, we can easily combine them
into the goal-oriented planning format: ``Goal: The target object is a \texttt{washing machine}. It is usually in a \texttt{laundry room}''.

\vspace{-8pt}
\subsubsection{Scene-Oriented Dynamic Planning (SODP)} 
\label{subsubsec:SODP}
\vspace{-4pt}
As is shown in Fig~\ref{fig:prompt}(b), the prompt for SODP consists of three parts.
The first part is based on the scene perception of room type, such as ``\texttt{bedroom}'', and visible objects, such as ``\texttt{bed, lamp, pillow}'', obtained by the Room-and-Object Aware Scene Perceiver (ROASP,
Sec.~\ref{subsec:monitor}).
These information are transformed into a natural language description of the current scene in the format of ``At this step, I am in \texttt{bedroom}, I can see \texttt{bed, lamp, pillow}''.
The second part is a demonstration of the fine-grained step-by-step instruction, which is selected according to the strategy detailed in the next section.
The last part is previous instructions, such as ``\textbf{Task}: Empty the washing machine on level one. 
 Step 1: ''.
All these three parts are concatenated together and then fed into an LLM to generate the fine-grained planning instruction for the next step accordingly, such as ``\texttt{Exit the bedroom}''. 

\vspace{-8pt}
\subsubsection{Dynamic Demonstration Selection}
\label{subsubsec:DDS}
\vspace{-4pt}
Recent works show that providing various demonstration examples to LLMs benefits the in-context learning for different tasks~\cite{huang2022language,liu2021makes,PoesiaP00SMG22}. In light of these findings, to direct the LLMs in generating better fine-grained plannings, we 
dynamically select the most suitable demonstration example for each specific task in REVERIE as the prompt to generate the environment-aware instruction, contrary to using a single fixed demonstration for all tasks.

\begin{figure}[t]
\begin{center}
\includegraphics[width=0.95\linewidth]{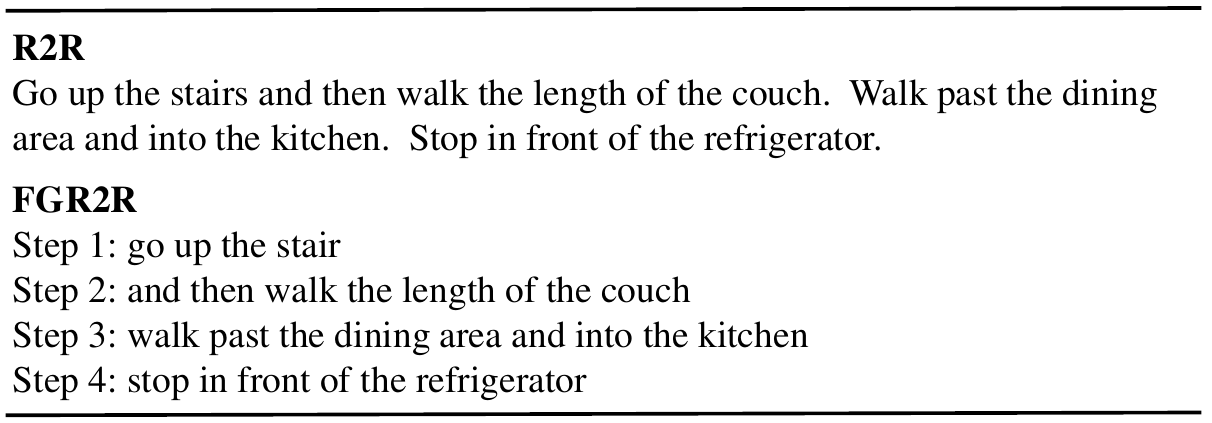}
\end{center}
\vspace{-15pt}
   \caption{Example of instructions in R2R and FGR2R.}
\vspace{-10pt}
    \label{fig:fgr2r}
\end{figure}

Specifically, we choose the training set of the Fine-Grained R2R dataset (FGR2R)~\cite{hong2020sub} as the demonstration set $\mathcal{D}$, of which each sample will be used as a demonstration example $D_{step}$. 
As shown in Fig.~\ref{fig:fgr2r}, FGR2R decomposes each low-level instruction $I_{low}$ of R2R dataset~\cite{r2r} into step-by-step instructions $I_{step}$.
Then, given a high-level instruction  $I_{high}$  of REVERIE, a proper $I_{low}$ will be selected as the demonstration example $D_{step}$ by a matching algorithm.
In particular, we use $I_{high}$ as query $Q$ and each low-level instruction $I_{low}$ as the key $K_i$, both of which are embedded by the Sentence-BERT~\cite{reimers2019sentence}. 
The semantic distance score between the two embeddings is calculated by the cosine similarity:
\begin{align}
    \text{s}(Q, K_i) = \frac{e(Q) \cdot e(K_i)}{\Vert e(Q) \Vert \Vert e(K_i) \Vert},
    \label{eq:cosine}
\end{align}
where $e(\cdot)$ is the embedding function. 
If $K_i$ has the highest similarity score to the given query $Q$, its corresponding step-by-step instruction $I_{step}$ will be selected as the demonstration example $D_{step}$ for the given high-level instruction $I_{high}$.

\subsection{Room-and-Object Aware Scene Perceiver}
\label{subsec:monitor}

Though the world knowledge acquired from the static LLMs planner could benefit the embodied task promisingly, the static LLMs planner may generate wrong or irrelevant plannings, which misleads the agent. 
To address this issue, the LLM planner should be aware of and interact with the dynamic observations. 
In~\cite{song2022llm}, the names of objects obtained from the ground truth or pre-trained detectors have been added to   in-context prompts. 
However, the agents of these works act in a very limited space with simple scenes and monotonous objects. 
By contrast,   REVERIE involves large-area exploration between different floors and rooms, where the scenes are more complex with more diverse objects.
Considering these factors, we propose a  room-and-object aware scene perceiver (ROASP) for the LLM planner, which   predicts not only the room type but also the visible objects of the current location. 
Rather than using separate classifiers and detectors to individually predict each position's room types and visible object categories,
we use CLIP~\cite{clip} as the proposed room-and-object aware scene perceiver. Thanks to CLIP's strong ability of zero-shot image classification in the open world, the ROASP can well handle these two tasks.

Specifically, we first fetch the room type labels from the MatterPort3D~\cite{mattport} semantic annotations
and the object type labels are extracted from the REVERIE training dataset. 
They are used to build the codebook for the room categories $\textbf{C}_\text{room}$ and the object categories $\textbf{C}_\text{obj}$, respectively.
Then, at each timestep $t$, the agent perceives the environment and obtains the 
panoramic visual observation $\mathcal R_t =\left \{ r_i \right \}_{i=1}^{n}$. 
For each single-view observation $r_i$ in the panorama, the image feature $f_\text{r}$ is extracted by the CLIP Image Encoder
\begin{equation}
    f_\text{r} =E_\text{CLIP}^{v} (r_i),
\end{equation}
where $E_\text{CLIP}^{v} (\cdot)$ represents the CLIP Image Encoder.
For each room category $c_{room}$ and each object category $c_{obj}$, we respectively construct a text phrase of room $T_\text{room}$ as ``a photo of a $\{c_{room}\}$'' and a text phrase of object $T_\text{obj}$ as ``a photo of a $\{c_{obj}\}$''.
Then the text feature is derived through the pretrained CLIP Text Encoder as:
\begin{align}
    f_{\text{room}} &= E_\text{CLIP}^{t} (T_\text{room}),\\
    f_{\text{obj}} &= E_\text{CLIP}^{t} (T_\text{obj}),
\end{align}
where $E_\text{CLIP}^{t}(\cdot)$ represents the CLIP Text Encoder.
At last, the similarity score ${S}_\text{room}$ between the image feature $f_\text{r}$ and the text feature $f_\text{room}$ 
as well as the similarity score ${S}_\text{obj}$ between the image feature $f_\text{r}$ and the text feature $f_\text{obj}$
are respectively computed as:
\begin{align}
    {S}_\text{room} &=\text{Softmax}(f_\text{room}\cdot f_\text{r}^{T}), \\
    {S}_\text{obj} &=\text{Softmax}(f_\text{obj}\cdot f_\text{r}^{T}). 
\end{align}

Considering that the current environment normally belongs to only one type of room, though the panoramic images have multiple views, the room that the agent is currently centered in should have the largest influence on each view. 
Thus, we average the predicted room type scores $S_\text{room}$ from multiple views and choose the room type with the greatest score as the room type prediction $\hat{c}_\text{room}$. 
For object predictions, if the object occupies more proportion in a view, the matching score ${S}_\text{obj}$ should be higher. Thus, we select $k$ prominent objects with the top-$k$ matching scores as the auxiliary environment feedback in addition to the predicted room.

\subsection{March with Interactive Prompting}
\label{subsec:march_with_inter_prompt}
When the generation of the goal-oriented planning and the scene-oriented planning with perceptions from the environment is finished, the agent can march towards the target object at each timestep $t$ under the guidance of the interactive prompting. In this section, we will give a detailed description of how the interactive prompting works during the process of navigation, which mainly consists of two parts, \ie the assembled instruction and the instruction update.

\begin{figure}[t]
\begin{center}
\includegraphics[width=0.99\linewidth]{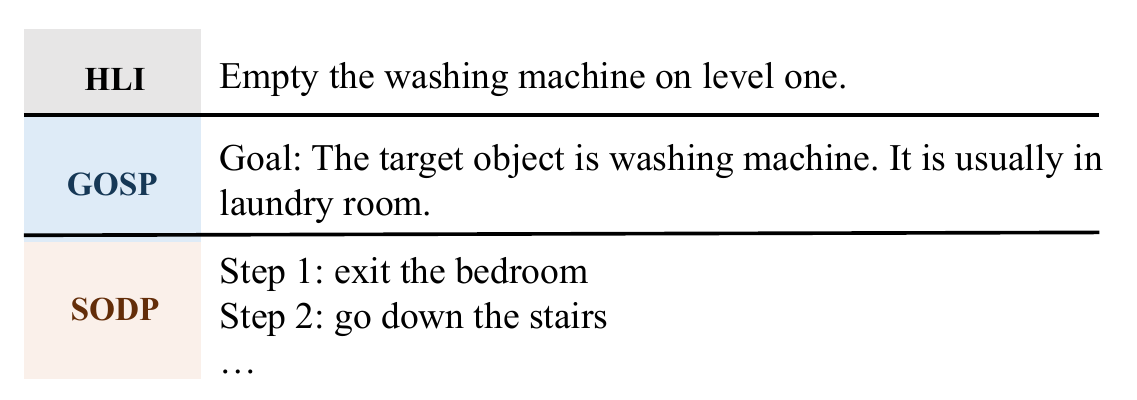}
\end{center}
\vspace{-14pt}
   \caption{Text inputs contains three parts: High-level Instruction in REVERIE (HLI), Goal-Oriented Static Planning (GOSP) and Scene-Oriented Dynamic Planning (SODP) returned instructions.}
\vspace{-10pt}
\label{fig:instruction}
\end{figure}

\vspace{-10pt}
\paragraph{Assembled Instruction}
\label{sec:inter_prompt}
At each timestep $t$, the agent observes the environment and receives the assembled instructions obtained from the above-mentioned modules, and choose an action to perform. Specifically, as shown in  Fig.~\ref{fig:instruction}, the assembled instructions $W$ of the interactive prompting mainly consist of three parts: the high-level instruction (HLI) $W_I$ in REVERIE, the GOSP instruction $W_G$ and the SODP instruction $W_S$.
We concatenate these three parts of instructions as the assembled instruction $W=[W_I, W_G, W_S]$ and use WordPieces~\cite{Wordpieces} to tokenize all the words into a sequence of tokens as the textual input for the agent. Then, the agent will act under the guidance of such assembled instruction. Note that the use of the original high-level instruction $W_I$ can improve the model's tolerance on the noise of intermediate planning instructions.

\vspace{-10pt}
\paragraph{Instruction Update}
The GOSP is only conducted once at the beginning of the task. While the SODP is conducted depending on the feedback of environments. Specifically, at each timestep $t$, if the ROASP finds the room has changed where the predicted room $\hat{c}_\text{room}^{t}$ does not equal to $\hat{c}_\text{room}^{t-1}$, the SODP will be triggered again. Then, a new step-by-step instruction such as ``Step 2: go down the stairs'' for the next few steps will be generated by the LLM and added to the previous assembled instruction $W$ after the last step-by-step instruction of ``Step 1: exit the bedroom''. Then, the agent will act under the guidance of the updated instructions $W'$.

\section{Experiment}
\label{sec:experiment}

\begin{table*}[!t]
\centering

\vspace{-0mm}
\resizebox{0.9\linewidth}{!}{
\begin{tabular}{l|cccccc|cccccc}
\toprule
\multicolumn{1}{l|}{\multirow{3}{*}{\textbf{Methods}}} &\multicolumn{6}{c|}{\textbf{Val Unseen}} & \multicolumn{6}{c}{\textbf{Test Unseen}} \\
~& \multicolumn{4}{c}{\textbf{Navigation}}  & \multicolumn{2}{c|}{\textbf{Grounding}}&\multicolumn{4}{c}{\textbf{Navigation}}   &\multicolumn{2}{c}{\textbf{Grounding}} \\
~& \multicolumn{1}{c}{TL} & \multicolumn{1}{c}{OSR$\uparrow$} & \multicolumn{1}{c}{SR$\uparrow$} & SPL$\uparrow$ &  \multicolumn{1}{c}{RGS$\uparrow$} & \multicolumn{1}{c|}{RGSPL$\uparrow$} & \multicolumn{1}{c}{TL} & 
\multicolumn{1}{c}{OSR$\uparrow$} & \multicolumn{1}{c}{SR$\uparrow$} & \multicolumn{1}{c}{\textbf{SPL}$\uparrow$}  &  \multicolumn{1}{c}{RGS$\uparrow$} & \multicolumn{1}{c}{RGSPL$\uparrow$} \\

\midrule
Human & -- & --  & --  & \cellcolor{c1}{--} & -- & \cellcolor{c1}-- & 21.18& 86.83 &81.51 & \cellcolor{c1}53.66 & 77.84 & \cellcolor{c1} 51.44 \\
\midrule
Seq2Seq &11.07&8.07&4.20&\cellcolor{c1}{2.84}&2.16&\cellcolor{c1}1.63&10.89&6.88&3.99&\cellcolor{c1}3.09&2.00&\cellcolor{c1}1.58\\
RCM \cite{rcm} &11.98&14.23& 9.29& \cellcolor{c1}6.97 & 4.89 & \cellcolor{c1}10.60 & 7.84 & 3.89 & 11.68 & \cellcolor{c1}6.67 & 3.67 & \cellcolor{c1}3.14\\
SMNA \cite{selfmonitor} & 9.07&11.28&8.15 &\cellcolor{c1}6.44 & 4.54&\cellcolor{c1} 3.61 & 9.23& 8.39&5.80& \cellcolor{c1}4.53 & 3.10& \cellcolor{c1}2.39 \\
FAST-MATTN \cite{reverie} & 45.28 & 28.20 & 14.40 & \cellcolor{c1}7.19  & 7.84 &\cellcolor{c1} 4.67  & 39.05  & {30.63} & 19.88& \cellcolor{c1}11.61& 11.28 & \cellcolor{c1}6.08 \\
ORIST~\cite{orist} & 10.90 &  25.02 & 16.84 &\cellcolor{c1}15.14 & 8.52 & \cellcolor{c1}7.58 & 11.38& 29.20 & 22.19 & \cellcolor{c1}18.97  & 10.68 & \cellcolor{c1}9.28\\
CKR~\cite{gao2021room}&26.26&31.44&19.14&\cellcolor{c1}11.84&11.45&\cellcolor{c1}-&22.46&30.40&22.00&\cellcolor{c1}14.25&11.60&\cellcolor{c1}-\\
RecBERT~\cite{recurrent}& 16.78& 35.02&30.67  & \cellcolor{c1}24.90  & 18.77 & \cellcolor{c1}15.27& 15.86 & 32.91  & 29.61 & \cellcolor{c1}23.99 & 16.50 & \cellcolor{c1}13.51 \\
Airbert~\cite{airbert} & 18.71 & 34.51 & 27.89 & \cellcolor{c1}21.88& 18.23 & \cellcolor{c1}14.18 & 17.91& 34.20 & 30.28 & \cellcolor{c1}23.61  & 16.83 & \cellcolor{c1}13.28 \\
HAMT~\cite{hamt}&14.08&36.84&32.95&\cellcolor{c1}30.20&18.92&\cellcolor{c1}17.28&13.62&33.41&30.40&\cellcolor{c1}26.67&14.88&\cellcolor{c1}12.08\\
HOP~\cite{hop}& 16.46 & 36.24 & 31.78 & \cellcolor{c1}26.11 & 18.85 &\cellcolor{c1}15.73 & 16.38 & 33.06& 30.17  &\cellcolor{c1}24.34 & 17.69 & \cellcolor{c1}14.34\\
TD-STP~\cite{ZhaoCGWYRX022}&-&39.48&34.88&\cellcolor{c1}27.32&21.16&\cellcolor{c1}16.56&-&40.26&35.89&\cellcolor{c1}27.51&19.88&\cellcolor{c1}15.40\\
DUET~\cite{Chen_2022_DUET} &22.11&51.07&46.98&\cellcolor{c1}33.73&32.15&\cellcolor{c1}23.03&21.30&56.91&52.51&\cellcolor{c1}36.06&31.88&\cellcolor{c1}22.06\\
HM3D-DUET~\cite{Chen_2022_HM3D_AutoVLN}&-&62.14&55.89&\cellcolor{c1}40.85&36.58&\cellcolor{c1}26.76&-&62.30&55.17&\cellcolor{c1}38.88&32.23&\cellcolor{c1}22.68\\
\midrule
MiC&20.64&\textbf{62.37}&\textbf{56.97}&\cellcolor{c1}\textbf{43.60}&\textbf{37.52}&\cellcolor{c1}\textbf{28.72}&\textbf{18.11}&\textbf{62.40}&\textbf{55.74}&\cellcolor{c1}\textbf{41.97}&\textbf{35.25}&\cellcolor{c1}\textbf{26.17}\\
\bottomrule
\end{tabular}
}
\caption{Comparison with the state-of-the-art methods on REVERIE.}
\label{tab:reverie}
\vspace{-3mm}
\end{table*}
\subsection{Evaluation Setup}
\label{subsec:eval_setup}
\paragraph{Dataset}
REVERIE~\cite{reverie} contains 10,567 panoramic images within 90 buildings (4,140 target objects divided into 489 categories) and 21,702 instructions with 18 words on average. Each target viewpoint has 7 distinct panoramic objects with 50 bounding boxes on average.
It consists of four splits: train, validation seen, validation unseen and test unseen. %

\vspace{-8pt}
\paragraph{Evaluation Metrics} The performance of agents is evaluated in two ways: navigation and object grounding. For the navigation sub-task, the metrics are \textbf{Success Rate (SR)}, \textbf{Oracle Success Rate (OSR)}, and \textbf{Success weighted by Path Length~\cite{spl} (SPL)}, where SPL is the main metric.
For the grounding sub-task, the metrics are \textbf{Remote Grounding Success rate (RGS)} and \textbf{RGS weighted by Path Length (RGSPL)}, where RGSPL is the main metric for this sub-task.
For all these metrics, higher is better.

\vspace{-8pt}
\begin{itemize}[noitemsep,topsep=2pt,leftmargin = 35pt]
\item[\textbf{\texttt{TL}}] \textbf{Trajectory Length} measures the average length of all the predicted navigation trajectories in meters.
\item[\textbf{\texttt{SR}}] \textbf{Success Rate} measures the ratio of successful tasks, of which the agent's stop location is less than 3 meters away from the target location.
\item[\textbf{\texttt{OSR}}] \textbf{Oracle Success Rate} measures the ratio of tasks of which one of its trajectory viewpoints can observe the target object within 3 meters. 
\item[\textbf{\texttt{SPL}}] \textbf{Success weighted by Path Length} trades-off SR (Success Rate) against TL (Trajectory Length). It measures both the accuracy and efficiency of navigation.
\item[\textbf{\texttt{RGS}}] \textbf{Remote Grounding Success rate} measures the ratio of tasks that successfully locate the target object.
\item[\textbf{\texttt{RGSPL}}] \textbf{RGS weighted by Path Length} is RGS. 
\end{itemize}

\paragraph{Implementation Details}
\label{sub:imple_details}
Our model is trained on a single 3090 GPU for 30,000 iterations. 
We set the batch size to 4 and the learning rate to $1\!\times\!10^{-5}$.  
The best model is selected according to performance on the validation unseen split.
We use the same pretrained model and augmented data as~\cite{Chen_2022_HM3D_AutoVLN} for a fair comparison.
For the LLMs, we use the public GPT-2~\cite{gpt2} model for in-context learning. 
For the scene preceptor, we keep the top 3 object predictions for each position.

\begin{table}[!t]
\centering
\resizebox{0.95\linewidth}{!}{
\begin{tabular}{l ccc cc}
\toprule
\multicolumn{1}{l}{\multirow{2}{*}{\textbf{Components}}}  & \multicolumn{3}{c}{\textbf{Navigation}}&\multicolumn{2}{c}{\textbf{Grounding}}\\
~ &\multicolumn{1}{c}{OSR$\uparrow$} & {SR$\uparrow$}&{SPL$\uparrow$}& RGS$\uparrow$ & RGSPL$\uparrow$\\
\midrule
\textbf{HLI(Baseline)}&58.02&52.71&40.49&34.93&26.82\\
\textbf{HLI+GOSP} &59.92&55.28&42.46&37.13&28.24\\
\textbf{HLI+SODP}& 60.72&56.26&42.94&36.80&27.81\\
\textbf{HLI+GOSP+SODP}&62.37&56.97&43.60&37.52&28.72\\
\bottomrule
\end{tabular}}
\caption{Ablation of different components in MiC. 
}
\vspace{-10pt}
\label{tab:abla_comp}
\end{table}

\subsection{Comparison with State-of-The-Art Methods}
\label{subsec:com_sota}
As shown in Table~\ref{tab:reverie}, we compare MiC with the state-of-the-art methods on the REVERIE benchmark. 
Our method outperforms previous methods in all metrics on both validation unseen and test unseen splits. 
Particularly, compared with the SoTA method HM3D-DUET~\cite{Chen_2022_HM3D_AutoVLN}, MiC outperforms HM3D-DUET by a large margin of 3.09\% in terms of the main navigation metric SPL and 3.49\% of the main object grounding metric RGSPL on the Test Unseen split.
Note that MiC shares the same pre-trained model with the HM3D-DUET, these promising result demonstrates that our method can effectively improve the navigation and object grounding ability of agents.

\begin{figure*}[t]
\begin{center}
\includegraphics[width=0.97\linewidth]{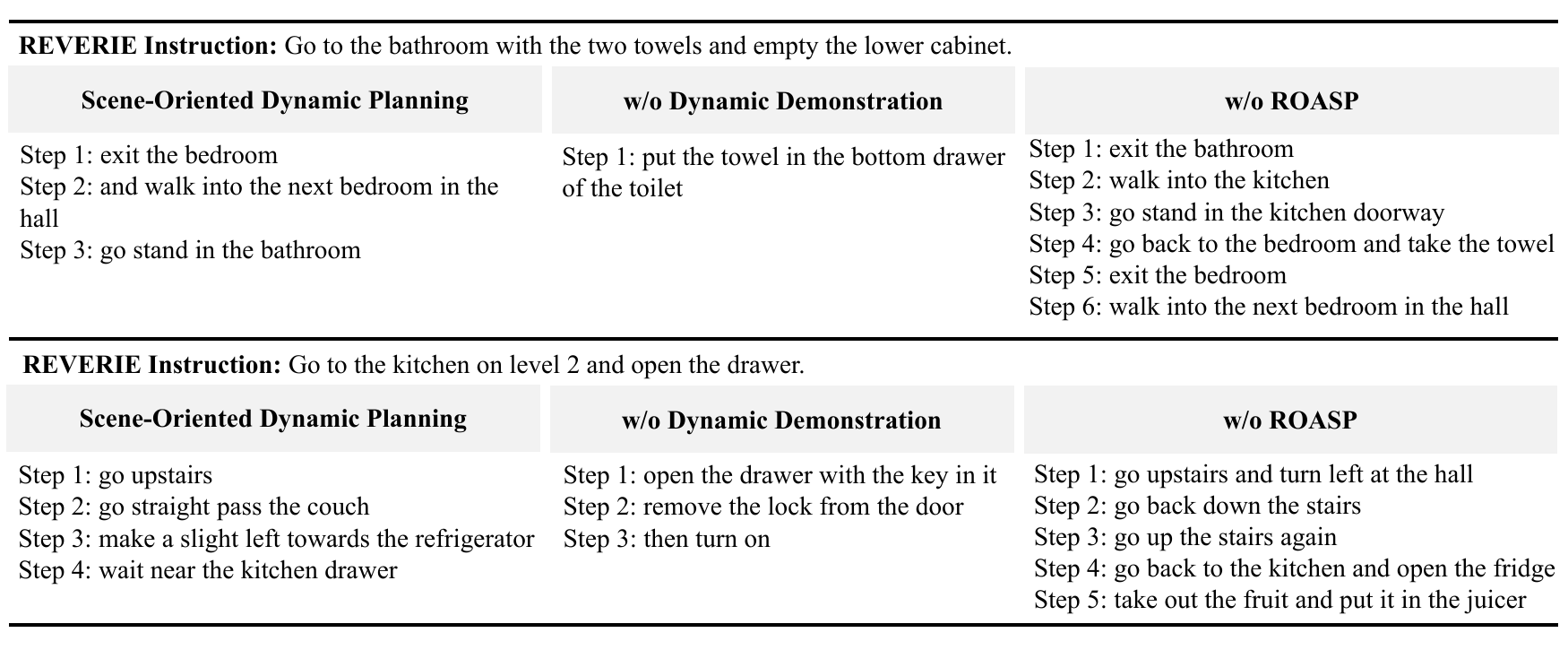}
\end{center}
\vspace{-15pt}
   \caption{Examples of generated instructions.}
\vspace{-15pt}
\label{fig:vis}
\end{figure*}

\subsection{Ablation Analysis}
\label{subsec:abla}
\paragraph{Contribution of different MiC Components}
In Table~\ref{tab:abla_comp}, we evaluate the effect of different components in our proposed MiC.
HLI denotes only using the original high-level instruction (HLI) provided by REVERIE.

Compared to the baseline HLI, GOSP improves the performance of both navigation (2.57\%$\uparrow$ on SR, 1.97\%$\uparrow$ on SPL) and object grounding (2.20\%$\uparrow$ on RGS, 1.42\%$\uparrow$ on RGSPL) with a non-trivial margin, showing the effectiveness of the proposed goal-oriented static planning. 
SODP further surpasses GOSP in the navigation metric (0.98\%$\uparrow$ on SR, 0.48\%$\uparrow$ on SPL) while falling a little behind in the grounding metrics (0.33\%$\uparrow$ on RGS, 1.43\%$\uparrow$ on RGSPL). 
The reason may be that the detailed step-by-step planning occupies a large proportion compared to the target object in the input texts, which  can bring the noise for object grounding while improving navigation performance.
When combining all these components, the final performance gets further increased in all metrics, which surpasses the baseline with a large margin (4.26\%$\uparrow$ on SR, 3.11\%$\uparrow$ on SPL, 2.59\%$\uparrow$ on RGS and 1.9\%$\uparrow$ on RGSPL). The promising results here show that these components are complementary to each other.

\begin{table}[!t]
\centering
\resizebox{0.9\linewidth}{!}{
\begin{tabular}{l cc ccc}
\toprule
\multicolumn{1}{l}{\multirow{2}{*}{\textbf{Methods}}} & \multicolumn{3}{c}{\textbf{Navigation}}&\multicolumn{2}{c}{\textbf{Grounding}}\\
~& \multicolumn{1}{c}{OSR$\uparrow$} & SR$\uparrow$& SPL$\uparrow$ & RGS$\uparrow$&RGSPL$\uparrow$\\
\midrule
\textbf{Baseline}&58.02&52.71&40.49&34.93&26.82\\
\textbf{Static}&60.24&55.35&41.74&36.30&27.03\\
\textbf{Dynamic}&60.72&56.26&42.94&36.80&27.81\\
\bottomrule
\end{tabular}}
\caption{Comparison of different plan generation settings. 
}
\label{tab:abla_static}
\end{table}

\begin{table}[!t]
\centering
\resizebox{0.99\linewidth}{!}{
\begin{tabular}{l cc}
\toprule
\textbf{Methods} &\textbf{Relevancy}& 
\textbf{Rationality}\\
\midrule
\textbf{Scene-Oriented Dynamic Planning}&2.06&1.93\\
- \textbf{w/o} Dynamic Demonstration&1.41&1.23\\
- \textbf{w/o} ROASP &1.64&1.55\\
\bottomrule
\end{tabular}}
\caption{Human study of the prompt setting for Scene-Oriented Dynamic Planning.
}
\vspace{-10pt}
\label{tab:abla_human}
\end{table}

\vspace{-8pt}
\paragraph{The Effect of the ROASP}
To evaluate the effectiveness of ROASP used for the scene-oriented dynamic planning, we conduct another ablation study via whether incorporating the feedback from the ROASP module on REVERIE validation unseen set. 
We report results in three settings: 
\textbf{(I)} Baseline: The input assembled instruction only contains the given high-level instruction in REVERIE.
\textbf{(II)} Static: The input assembled instruction contains the REVERIE and fine-grained static instructions. The difference between fine-grained static instruction and scene-oriented dynamic instruction is that static fine-grained instruction is generated without ROASP. More specifically, the query prompt for the LLM to generate step-by-step planning is fixed at each timestep, which only consists of the given high-level instruction and the selected demonstration.
\textbf{(III)} Dynamic: The input assembled instruction contains the high-level instruction in REVERIE and scene-oriented dynamic planning instruction. As shown in Table~\ref{tab:abla_static}, in the static setting, the performance in all metrics is improved compared to the baseline, indicating the effectiveness of the LLM's rich world knowledge in fine-grained planning. In the dynamic setting, the performance is further improved with non-trivial margins, showing the effectiveness of ROASP.

\paragraph{Qualitative Analysis of Prompt Setting}
To further evaluate the effect of dynamic demonstration and ROASP in SODP, we perform a human evaluation about the generated plannings (see Table~\ref{tab:abla_human}) and show the planning results (see Fig.~\ref{fig:vis}).
For human evaluation, we randomly selected 100 REVERIE tasks and generate fine-grained step-by-step instructions in setting of SODP, SODP without dynamic demonstration, and SODP without ROASP.
We asked 10 volunteers to mark the generated step-by-step instructions in terms of their relevancy and rationality.
The relevancy score ranges from 0 (unrelated) to 3 (very related), which takes into account whether the keywords in instructions are related to the REVERIE task. 
For example, regarding the REVERIE instruction   ``Go to the kitchen and turn on the microwave'', whether there are keywords in instructions related to the kitchen scene could be rated. Rationality is rated from 0 (bad) to 3 (perfect), considering whether the instruction conforms to the logic of navigation.

The results are presented in Table~\ref{tab:abla_human}. It shows that our SODP scored 2.06 on Relevancy and 1.93 on Rationality, which could be considered acceptable since the highest score is 3 and it is challenging to generate instructions that are consistent with tasks and actual navigation logic.
When removing the dynamic demonstration, the score of generated instruction drops about 31.55\% on Relevancy and 36.27\% on Rationality, which could also be observed in Fig.~\ref{fig:vis}. 
Although the instruction generated without dynamic demonstration is related to the task to some extent (\eg, ``put the towel in the bottom drawer of the toilet'' has the keyword ``towel'', the instruction lacks navigation information, such as how to reach the bathroom.)
As shown in the bottom example of Fig.~\ref{fig:vis}, instruction without ROASP successfully guided how to go to the destination location kitchen, but it still caused confusion by going upstairs and going downstairs several times, and thus reducing the rationality score, \ie 1.64 on Relevancy and 1.55 on Rationality. More generation results can be found in the  supplementary.

\section{Conclusions}
In this work, we propose a novel model, March-in-Chat (MiC), for the REVEIRE task, which only provides concise high-level instructions for the VLN agent. MiC enables the REVERIE agent to talk with an LLM on the fly to generate plans for the next few steps. It consists of three main  modules,  Goal-Oriented Static Planning (GOSP), Scene-Oriented Dynamic Planning (SODP), and  Room-and-Object Aware Scene Perceiver (ROASP) module. We conduct extensive quantitative and qualitative experiments on REVERIE and the promising results show the effectiveness of our method.

\section{Ackowledgements}
Jing Liu is supported by the National Key Research and Development Program of China (No. 2020AAA0106400).

{\small
\bibliographystyle{ieee_fullname}
\bibliography{egbib}

\begin{thebibliography}{10}\itemsep=-1pt

\bibitem{saycan}
Michael Ahn, Anthony Brohan, Noah Brown, Yevgen Chebotar, Omar Cortes, Byron
  David, Chelsea Finn, Chuyuan Fu, Keerthana Gopalakrishnan, Karol Hausman,
  Alex Herzog, Daniel Ho, Jasmine Hsu, Julian Ibarz, Brian Ichter, Alex Irpan,
  Eric Jang, Rosario~Jauregui Ruano, Kyle Jeffrey, Sally Jesmonth, Nikhil
  Joshi, Ryan Julian, Dmitry Kalashnikov, Yuheng Kuang, Kuang-Huei Lee, Sergey
  Levine, Yao Lu, Linda Luu, Carolina Parada, Peter Pastor, Jornell Quiambao,
  Kanishka Rao, Jarek Rettinghouse, Diego Reyes, Pierre Sermanet, Nicolas
  Sievers, Clayton Tan, Alexander Toshev, Vincent Vanhoucke, Fei Xia, Ted Xiao,
  Peng Xu, Sichun Xu, Mengyuan Yan, and Andy Zeng.
\newblock Do as i can and not as i say: Grounding language in robotic
  affordances.
\newblock In {\em arXiv preprint arXiv:2204.01691}, 2022.

\bibitem{an2022bevbert}
Dong An, Yuankai Qi, Yangguang Li, Yan Huang, Liang Wang, Tieniu Tan, and Jing
  Shao.
\newblock Bevbert: Topo-metric map pre-training for language-guided navigation.
\newblock {\em arXiv preprint arXiv:2212.04385}, 2022.

\bibitem{spl}
Peter Anderson, Angel~X. Chang, Devendra~Singh Chaplot, Alexey Dosovitskiy,
  Saurabh Gupta, Vladlen Koltun, Jana Kosecka, Jitendra Malik, Roozbeh
  Mottaghi, Manolis Savva, and Amir~Roshan Zamir.
\newblock On evaluation of embodied navigation agents.
\newblock {\em CoRR}, abs/1807.06757, 2018.

\bibitem{r2r}
Peter Anderson, Qi Wu, Damien Teney, Jake Bruce, Mark Johnson, Niko
  S{\"{u}}nderhauf, Ian~D. Reid, Stephen Gould, and Anton van~den Hengel.
\newblock Vision-and-language navigation: Interpreting visually-grounded
  navigation instructions in real environments.
\newblock In {\em CVPR}, pages 3674--3683, 2018.

\bibitem{gpt3}
Tom Brown, Benjamin Mann, Nick Ryder, Melanie Subbiah, Jared~D Kaplan, Prafulla
  Dhariwal, Arvind Neelakantan, Pranav Shyam, Girish Sastry, Amanda Askell,
  Sandhini Agarwal, Ariel Herbert-Voss, Gretchen Krueger, Tom Henighan, Rewon
  Child, Aditya Ramesh, Daniel Ziegler, Jeffrey Wu, Clemens Winter, Chris
  Hesse, Mark Chen, Eric Sigler, Mateusz Litwin, Scott Gray, Benjamin Chess,
  Jack Clark, Christopher Berner, Sam McCandlish, Alec Radford, Ilya Sutskever,
  and Dario Amodei.
\newblock Language models are few-shot learners.
\newblock In {\em NeurIPS}, pages 1877--1901, 2020.

\bibitem{mattport}
Angel~X. Chang, Angela Dai, Thomas~A. Funkhouser, Maciej Halber, Matthias
  Nie{\ss}ner, Manolis Savva, Shuran Song, Andy Zeng, and Yinda Zhang.
\newblock Matterport3d: Learning from {RGB-D} data in indoor environments.
\newblock In {\em 3DV}, pages 667--676, 2017.

\bibitem{touchdown}
Howard Chen, Alane Suhr, Dipendra Misra, Noah Snavely, and Yoav Artzi.
\newblock {TOUCHDOWN:} natural language navigation and spatial reasoning in
  visual street environments.
\newblock In {\em CVPR}, pages 12538--12547, 2019.

\bibitem{codex}
Mark Chen, Jerry Tworek, Heewoo Jun, Qiming Yuan, Henrique Ponde de~Oliveira
  Pinto, Jared Kaplan, Harri Edwards, Yuri Burda, Nicholas Joseph, Greg
  Brockman, et~al.
\newblock Evaluating large language models trained on code.
\newblock {\em arXiv preprint arXiv:2107.03374}, 2021.

\bibitem{hamt}
Shizhe Chen, Pierre{-}Louis Guhur, Cordelia Schmid, and Ivan Laptev.
\newblock History aware multimodal transformer for vision-and-language
  navigation.
\newblock In {\em NeurIPS}, pages 5834--5847, 2021.

\bibitem{Chen_2022_HM3D_AutoVLN}
Shizhe Chen, Pierre-Louis Guhur, Makarand Tapaswi, Cordelia Schmid, and Ivan
  Laptev.
\newblock Learning from unlabeled 3d environments for vision-and-language
  navigation.
\newblock In {\em ECCV}, 2022.

\bibitem{Chen_2022_DUET}
Shizhe Chen, Pierre-Louis Guhur, Makarand Tapaswi, Cordelia Schmid, and Ivan
  Laptev.
\newblock Think global, act local: Dual-scale graph transformer for
  vision-and-language navigation.
\newblock In {\em CVPR}, 2022.

\bibitem{gao2021room}
Chen Gao, Jinyu Chen, Si Liu, Luting Wang, Qiong Zhang, and Qi Wu.
\newblock Room-and-object aware knowledge reasoning for remote embodied
  referring expression.
\newblock In {\em CVPR}, pages 3064--3073, 2021.

\bibitem{airbert}
Pierre{-}Louis Guhur, Makarand Tapaswi, Shizhe Chen, Ivan Laptev, and Cordelia
  Schmid.
\newblock Airbert: In-domain pretraining for vision-and-language navigation.
\newblock In {\em ICCV}, pages 1634--1643, 2021.

\bibitem{he2021landmark}
Keji He, Yan Huang, Qi Wu, Jianhua Yang, Dong An, Shuanglin Sima, and Liang
  Wang.
\newblock Landmark-rxr: Solving vision-and-language navigation with
  fine-grained alignment supervision.
\newblock {\em NeurIPS}, 34:652--663, 2021.

\bibitem{hong2020sub}
Yicong Hong, Cristian~Rodriguez Opazo, Qi Wu, and Stephen Gould.
\newblock Sub-instruction aware vision-and-language navigation.
\newblock In Bonnie Webber, Trevor Cohn, Yulan He, and Yang Liu, editors, {\em
  {EMNLP}}, pages 3360--3376, 2020.

\bibitem{recurrent}
Yicong Hong, Qi Wu, Yuankai Qi, Cristian~Rodriguez Opazo, and Stephen Gould.
\newblock {\rvlnbert:} {A} recurrent vision-and-language {BERT} for navigation.
\newblock In {\em CVPR}, pages 1643--1653, 2021.

\bibitem{huang2022language}
Wenlong Huang, Pieter Abbeel, Deepak Pathak, and Igor Mordatch.
\newblock Language models as zero-shot planners: Extracting actionable
  knowledge for embodied agents.
\newblock In {\em ICML}, pages 9118--9147. PMLR, 2022.

\bibitem{huang2022inner}
Wenlong Huang, Fei Xia, Ted Xiao, Harris Chan, Jacky Liang, Pete Florence, Andy
  Zeng, Jonathan Tompson, Igor Mordatch, Yevgen Chebotar, Pierre Sermanet, Noah
  Brown, Tomas Jackson, Linda Luu, Sergey Levine, Karol Hausman, and Brian
  Ichter.
\newblock Inner monologue: Embodied reasoning through planning with language
  models.
\newblock In {\em arXiv preprint arXiv:2207.05608}, 2022.

\bibitem{r4r}
Vihan Jain, Gabriel Magalh{\~{a}}es, Alexander Ku, Ashish Vaswani, Eugene Ie,
  and Jason Baldridge.
\newblock Stay on the path: Instruction fidelity in vision-and-language
  navigation.
\newblock pages 1862--1872, 2019.

\bibitem{Wordpieces}
Melvin Johnson, Mike Schuster, Quoc~V. Le, Maxim Krikun, Yonghui Wu, Zhifeng
  Chen, Nikhil Thorat, Fernanda~B. Vi{\'{e}}gas, Martin Wattenberg, Greg
  Corrado, Macduff Hughes, and Jeffrey Dean.
\newblock Google's multilingual neural machine translation system: Enabling
  zero-shot translation.
\newblock {\em Trans. Assoc. Comput. Linguistics}, 5:339--351, 2017.

\bibitem{rxr}
Alexander Ku, Peter Anderson, Roma Patel, Eugene Ie, and Jason Baldridge.
\newblock Room-across-room: Multilingual vision-and-language navigation with
  dense spatiotemporal grounding.
\newblock In {\em EMNLP}, pages 4392--4412, 2020.

\bibitem{liu2021makes}
Jiachang Liu, Dinghan Shen, Yizhe Zhang, Bill Dolan, Lawrence Carin, and Weizhu
  Chen.
\newblock What makes good in-context examples for gpt-$3$?
\newblock {\em arXiv preprint arXiv:2101.06804}, 2021.

\bibitem{selfmonitor}
Chih{-}Yao Ma, Jiasen Lu, Zuxuan Wu, Ghassan AlRegib, Zsolt Kira, Richard
  Socher, and Caiming Xiong.
\newblock Self-monitoring navigation agent via auxiliary progress estimation.
\newblock In {\em ICLR}, 2019.

\bibitem{hanna}
Khanh Nguyen and Hal~Daum{\'{e}} III.
\newblock Help, anna! visual navigation with natural multimodal assistance via
  retrospective curiosity-encouraging imitation learning.
\newblock pages 684--695. Association for Computational Linguistics, 2019.

\bibitem{PoesiaP00SMG22}
Gabriel Poesia, Alex Polozov, Vu Le, Ashish Tiwari, Gustavo Soares, Christopher
  Meek, and Sumit Gulwani.
\newblock Synchromesh: Reliable code generation from pre-trained language
  models.
\newblock In {\em ICLR}, 2022.

\bibitem{orist}
Yuankai Qi, Zizheng Pan, Yicong Hong, Ming{-}Hsuan Yang, Anton van~den Hengel,
  and Qi Wu.
\newblock The road to know-where: An object-and-room informed sequential bert
  for indoor vision-language navigation.
\newblock In {\em ICCV}, pages 1655--1664, 2021.

\bibitem{reverie}
Yuankai Qi, Qi Wu, Peter Anderson, Xin Wang, William~Yang Wang, Chunhua Shen,
  and Anton van~den Hengel.
\newblock {REVERIE:} remote embodied visual referring expression in real indoor
  environments.
\newblock In {\em CVPR}, pages 9979--9988, 2020.

\bibitem{hop}
Yanyuan Qiao, Yuankai Qi, Yicong Hong, Zheng Yu, Peng Wang, and Qi Wu.
\newblock {HOP:} history-and-order aware pre-training for vision-and-language
  navigation.
\newblock In {\em CVPR}, pages 15397--15406, 2022.

\bibitem{qiao2023hop+}
Yanyuan Qiao, Yuankai Qi, Yicong Hong, Zheng Yu, Peng Wang, and Qi Wu.
\newblock Hop+: History-enhanced and order-aware pre-training for
  vision-and-language navigation.
\newblock {\em IEEE TPAMI}, 2023.

\bibitem{clip}
Alec Radford, Jong~Wook Kim, Chris Hallacy, Aditya Ramesh, Gabriel Goh,
  Sandhini Agarwal, Girish Sastry, Amanda Askell, Pamela Mishkin, Jack Clark,
  Gretchen Krueger, and Ilya Sutskever.
\newblock Learning transferable visual models from natural language
  supervision.
\newblock In {\em ICML}, volume 139, pages 8748--8763, 2021.

\bibitem{gpt2}
Alec Radford, Jeffrey Wu, Rewon Child, David Luan, Dario Amodei, Ilya
  Sutskever, et~al.
\newblock Language models are unsupervised multitask learners.
\newblock {\em OpenAI blog}, 1(8):9, 2019.

\bibitem{reimers2019sentence}
Nils Reimers and Iryna Gurevych.
\newblock Sentence-bert: Sentence embeddings using siamese bert-networks.
\newblock In Kentaro Inui, Jing Jiang, Vincent Ng, and Xiaojun Wan, editors,
  {\em EMNLP-IJCNLP}, pages 3980--3990, 2019.

\bibitem{alfred}
Mohit Shridhar, Jesse Thomason, Daniel Gordon, Yonatan Bisk, Winson Han,
  Roozbeh Mottaghi, Luke Zettlemoyer, and Dieter Fox.
\newblock {ALFRED:} {A} benchmark for interpreting grounded instructions for
  everyday tasks.
\newblock In {\em CVPR}, pages 10737--10746, 2020.

\bibitem{progprompt}
Ishika Singh, Valts Blukis, Arsalan Mousavian, Ankit Goyal, Danfei Xu, Jonathan
  Tremblay, Dieter Fox, Jesse Thomason, and Animesh Garg.
\newblock {ProgPrompt}: Generating situated robot task plans using large
  language models.
\newblock 2022.

\bibitem{song2022llm}
Chan~Hee Song, Jiaman Wu, Clayton Washington, Brian~M Sadler, Wei-Lun Chao, and
  Yu Su.
\newblock Llm-planner: Few-shot grounded planning for embodied agents with
  large language models.
\newblock {\em arXiv preprint arXiv:2212.04088}, 2022.

\bibitem{ndh}
Jesse Thomason, Michael Murray, Maya Cakmak, and Luke Zettlemoyer.
\newblock Vision-and-dialog navigation.
\newblock In {\em CoRL}, pages 394--406, 2019.

\bibitem{rcm}
Xin Wang, Qiuyuan Huang, Asli {\c{C}}elikyilmaz, Jianfeng Gao, Dinghan Shen,
  Yuan{-}Fang Wang, William~Yang Wang, and Lei Zhang.
\newblock Reinforced cross-modal matching and self-supervised imitation
  learning for vision-language navigation.
\newblock In {\em CVPR}, pages 6629--6638, 2019.

\bibitem{ZhaoCGWYRX022}
Yusheng Zhao, Jinyu Chen, Chen Gao, Wenguan Wang, Lirong Yang, Haibing Ren,
  Huaxia Xia, and Si Liu.
\newblock Target-driven structured transformer planner for vision-language
  navigation.
\newblock pages 4194--4203, 2022.

\bibitem{soon}
Fengda Zhu, Xiwen Liang, Yi Zhu, Qizhi Yu, Xiaojun Chang, and Xiaodan Liang.
\newblock {SOON:} scenario oriented object navigation with graph-based
  exploration.
\newblock In {\em CVPR}, pages 12689--12699, 2021.

\end{thebibliography}
}

\end{document}